\documentclass[11pt]{article}
\usepackage{natbib}
\DeclareUnicodeCharacter{1E24}{\d{H}}

\usepackage[preprint]{acl}
\usepackage{booktabs}
\usepackage{multirow}
\usepackage{times}
\usepackage{latexsym}
\usepackage{amsmath}
\usepackage[T1]{fontenc}
\usepackage[utf8]{inputenc}

\usepackage{microtype}

\usepackage{inconsolata}

\usepackage{graphicx}

\usepackage{booktabs}   % For professional-looking tables
\usepackage{caption}    % Better caption control
\usepackage{array}      % For column formatting
\usepackage{ragged2e}   % Optional: better text alignment

\title{AlignAR: Generative Sentence Alignment for Arabic–English Parallel Corpora of Legal and Literary Texts}

\author{
 \textbf{Baorong Huang\textsuperscript{1}},
 \textbf{Ali Asiri\textsuperscript{2}\thanks{Corresponding author: \href{mailto:amaasiri@uqu.edu.sa}{amaasiri@uqu.edu.sa}}},
\\
 \textsuperscript{1}School of Foreign Languages, Huaihua University
 \textsuperscript{2}Al-lith University College, Umm al-Qura University
\\
}

\begin{document}

\maketitle
% \thank{*Corresponding author: \href{mailto:amaasiri@uqu.edu.sa}{amaasiri@uqu.edu.sa}}
\begin{abstract}
High-quality parallel corpora are essential for Machine Translation (MT) research and translation teaching. However, Arabic–English resources remain scarce and existing datasets mainly consist of simple one-to-one mappings. In this paper, we present AlignAR, a generative sentence alignment method, and a new Arabic–English dataset comprising simple legal and complex literary parallel texts. Our evaluation demonstrates that  "Easy" datasets lack the discriminatory power to fully assess alignment methods. By reducing one-to-one mappings in our "Hard" subset, we exposed the limitations of traditional alignment methods. In contrast, LLM-based approaches demonstrated better robustness, achieving an overall F1-score of 85.5\%, a nearly 9\% improvement over previous methods. Our datasets and codes are open-sourced at https://github.com/XXX.

\end{abstract}

\section{Introduction}

Parallel sentence alignment is an important step in preparing parallel data for various Natural Language Processing (NLP) applications, particularly machine translation. It involves identifying corresponding sentences between two texts that are mutual translations, as shown in Figure \ref{fig:alignexample}. Given that manual alignment is resource-intensive and time-consuming, various automatic methods are proposed to address the challenges. 
\setlength{\abovecaptionskip}{10pt}  % default is ~10pt
\begin{figure}[htbp]
  \centering
  \includegraphics[width=1.0\linewidth]{latex/images/alignexample.png}
  \caption{Examples of 1-to-1, 1-to-many and many-to-many alignments between source and target sentences written in Arabic and English, respectively.}
  \label{fig:alignexample}
\end{figure}

Over the past three decades, the research has progressed from early statistical and length-based models \cite{Brown1991, GaleChurch1993} to heuristic, structural, and modern neural embedding-based approaches capable of handling diverse domains, combining multilingual sentence embeddings, anchor-point detection, and dynamic alignment algorithms, such as dictionary-based extraction\cite{Althobaiti2022}, divide and conquer\cite{steingrimsson2023sentalign, zhang2022improve}. More recent work explores embedding-based methods, such as overlapping fixed-length segmentation\cite{wang2024document},  multilingual embeddings \cite{thompson2019vecalign, liu2023bertalign}, contextualized document-aware sentence embeddings\cite{molfese2024crocoalign}, and particle swarm optimization\cite{ShangLi2024}. Despite these advances, research on sentence alignment using Large Language Models (LLMs) remains notably scarce. 

Previous alignment methods are evaluated on web datasets in English-German texts\cite{koehn2018findings} or some low-resource languages, such as Pashto–English\cite{koehn2020findings} or Estonian-Lithuanian\cite{sloto2023findings}. In addition, some gold alignment datasets are released to evaluate the alignment methods for French-German pairs such as Text+Berg, Chinese-English pairs\cite{liu2023bertalign},  English–Icelandic pairs\cite{steingrimsson2023sentalign}, or various European languages\cite{molfese2024crocoalign}. However, the resources on English-Arabic pairs are underexplored.

% Full-width table using table*
\begin{table*}[t]
\setlength{\tabcolsep}{10pt}  % default is 6pt
\centering
\caption{Statistics of the Arabic--English Parallel Corpus}
\label{tab:corpus_stats}
\begin{tabular}{lcccccc}
\toprule
Dataset & \#AR-SENT & \#AR-TKN & \#EN-SENT & \#EN-TKN & SENT. (\%) & 1-1 (\%) \\
\midrule
Easy 1 &  153 & 4,405    &   206 & 7,783    &   74.27       & 78.15      \\
Easy 2 & 202  &  6,437   &  257  & 11,438    &   78.60       &  77.78     \\
Easy 3 & 161  & 4,726    &  198  & 8,471    &  81.31        &  81.76     \\
Easy 4 & 202  & 4,893    & 233   & 8,766    &   87.73       &  85.20     \\
Easy 5 & 174  &  5,770   & 199   & 10,650    &   87.44       &  87.21     \\
Hard 1 & 93 & 3,880 & 202 & 6,318 & 46.03 & 28.77 \\
Hard 2 & 100 & 3,998 & 194 & 6,823 & 51.55 & 40.26 \\
Hard 3 & 93 & 3,111 & 175 & 5,849 & 53.14 & 45.78 \\
Hard 4 & 92 & 3,068 & 203 & 5,800 & 45.32 & 40.70 \\
Hard 5 & 101 & 2,759 & 223 & 6,906 & 45.29 & 31.18 \\
\bottomrule
\end{tabular}

\medskip
\justify
\footnotesize
\noindent
\textbf{Note:} \#AR-SENT and \#AR-TKN indicate sentence count and token count in the Arabic text; \#EN-SENT and \#EN-TKN indicate sentence count and token count in the English text; 1-1 (\%) means the frequency (percentage) of 1-to-1 alignments in each dataset. Both source and target texts have been tokenized with Stanza before counting tokens.
\end{table*}

To address these limitations, we conduct an in-depth investigation into the integration of Large Language Models (LLMs) for automated sentence alignment and put forward the following two main contributions:
\begin{itemize}
\item We introduce a novel LLM-based alignment method. This method leverages the contextual understanding of LLMs to generate high-precision initial alignments, which we rigorously evaluate against our specialized parallel corpus;

\item We present the Arabic–English Parallel Corpus, a new gold-standard dataset designed to test alignment robustness across varying levels of difficulty. The corpus is categorized into a legal sub-corpus, representing structured technical prose, and a literary sub-corpus, containing complex, non-linear narratives that pose a significant challenge to previous alignment methods.
\end{itemize}
Ultimately, we hope that this study provides both a scalable methodology and a high-fidelity resource to stimulate further research into specialized bilingual alignment, particularly for low-resource and linguistically complex pairs.

\section{Existing methods}
\subsection{BleuAlign}
\citet{sennrich2010mt} proposed Bleualign, in which high-scoring sentence pairs are treated as anchor points, while dynamic programming is employed to infer optimal alignments between anchors, allowing for one-to-one and one-to-many mappings. The approach also integrates traditional length-based heuristics as a fallback mechanism in difficult cases. 
\subsection{VecAlign}
\citet{thompson2019vecalign} introduced Vecalign, which represents sentences from both languages in a shared semantic vector space using pre-trained multilingual encoders and computes cosine similarity to identify semantically corresponding sentence pairs. To ensure scalability, the authors proposed a linear-time dynamic programming algorithm that aligns sentences monotonically while permitting sentence splits and merges. 
\subsection{BertAlign}
\citet{liu2023bertalign} proposed Bertalign, in which sentences are encoded using transformer embeddings, enabling more precise similarity estimation than static word or sentence embeddings. Bertalign computes sentence-level similarity scores and applies dynamic programming to determine optimal alignments under monotonicity constraints.

\begin{figure*}[htbp]
  \centering
  \includegraphics[width=1.0\linewidth]{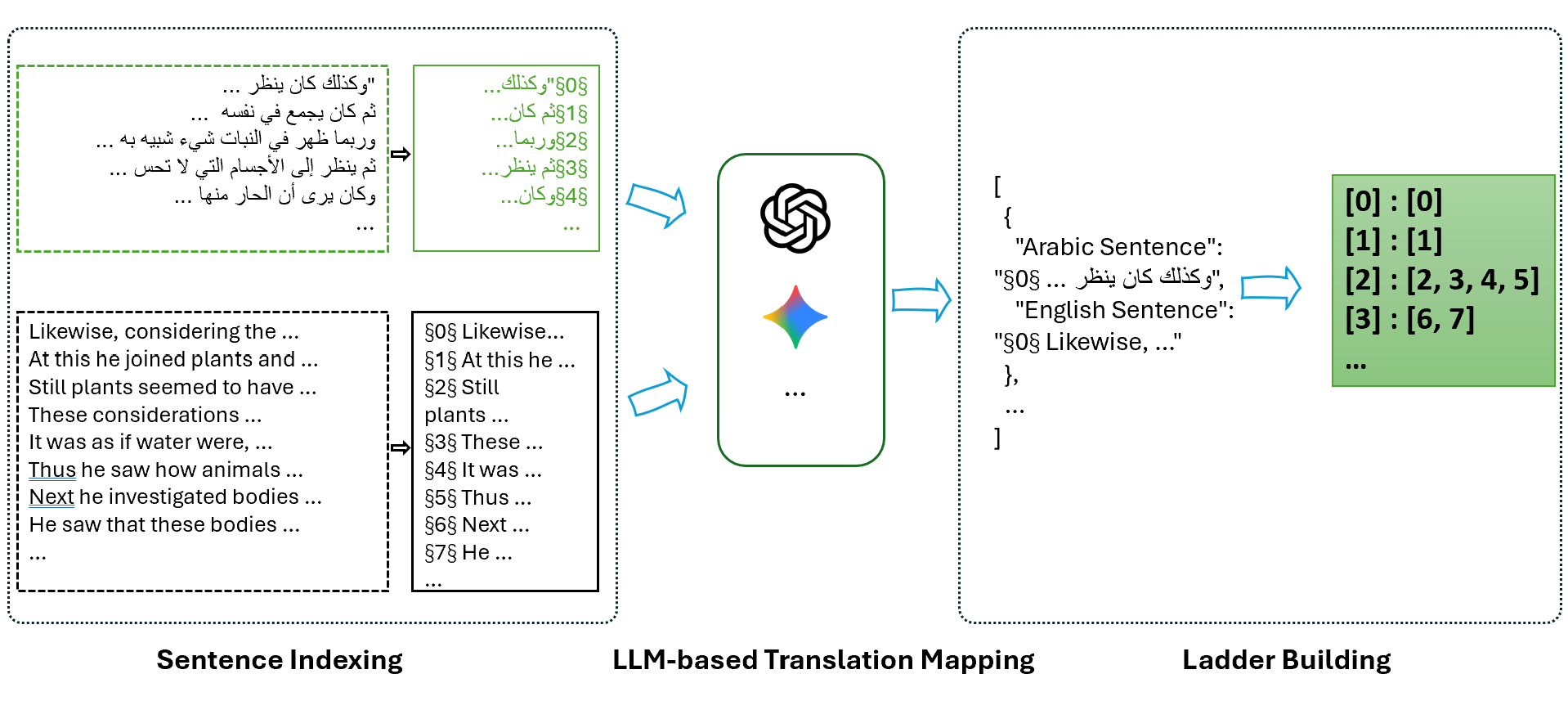}
  \caption{Our alignment procedure consists of three steps: (1) sentence indexing, i.e., explicitly indexing the lines with sequence number; (2) LLM-based translation mapping, i.e., requesting the selected LLM to select the translations from the target lines; (3) ladder building, extracting the index number from the formatted JSON output and obtaining ladder files for evaluation.}
  \label{fig:procedure}
\end{figure*}

\section{Datasets}
The data used in this study are divided into two main categories based on the complexity of the sentence alignment task:  easy-alignment data and hard-alignment data, as shown in Table ~\ref{tab:corpus_stats}. the hard part consists of 5 documents, totaling 378 Arabic sentences and 774 English sentences, with the smallest source/target ratio around 0.45, which means that one Arabic sentence aligns to at least two English sentences on average. The easy part also consists of 5 documents manually aligned from Arabic laws and its English translations, totaling 892 source lines and 1093 target lines, with the smallest source/target ratio around 0.74, much higher than the hard part. 

The easy-alignment subset comprises modern legal texts characterized by high structural parallelism and terminological consistency. These data were collected from the Saudi National Center for Archives and Records (NCAR), which provides official Saudi laws alongside their English translations, including the Companies Law, Anti-Bribery Law, and Anti-Commercial Concealment Law. Due to their formulaic drafting style, controlled vocabulary, and close translation correspondence, these legal texts represent low-complexity alignment data.

In contrast, the data of the hard-alignment subset are drawn from classical and literary Arabic texts, which present significant challenges due to archaic language, stylistic density, and non-literal translation strategies. The raw data are from the open-source platform Rasaifa, including the philosophical narrative \d{H}ayyibn Yaq\d{z}\=an, and short stories included in Hassan Ghazala’s A Textbook of Literary Translation (2013).

\section{Method}
In this study, we frame document alignment as a zero-shot inference task, prompting the models to identify correspondences based on semantic equivalence rather than lexical overlap. We propose a novel sentence alignment that considers alignment as a translation mapping task, and request the LLM to identify the translations of the source text, rather than directly performing the alignment.
\subsection{Alignment Procedure}
We propose a three-step alignment procedure. In the first step, we intuitively add a sentence index (sequence number) as [$0\ s_0,\ \dots,\ n\ s_m$] and [$0\ t_0,\ \dots,\ n\ t_n$] to each source line and each target line to reduce the hallucinations produced by an LLM. To further reduce the inadvertent addition or deletion of words by an LLM, we change the parallel aligning task into a translation mapping task, requesting the LLM to pick out the corresponding translations of the source lines from the target lines and output the entire mappings in a JSON format. In the final step, we extract the sentence index to obtain a ladder file that only contains line number mappings for evaluation, as shown in Figure \ref{fig:procedure}.  
% We used the openly accessible API of the large language models (LLMs), including GPT and Gemini, in the document alignments. To reduce the hallucinations inherent to LLMs, the lines in the document are first labeled with line numbers as [$0\ s_0,\ \dots,\ n\ s_m$] and [$0\ t_0,\ \dots,\ n\ t_n$], and then encapsulated into the prompt (see Appendix) that requested the large language models to spot the translation relations between the source document and target document, returning the response as a translation mapping from source lines to target lines. The resulting responses are then processed to , as shown in Figure \ref{fig:procedure}.

\begin{table}[htbp]
\setlength{\tabcolsep}{2pt}  % default is 6pt
\centering
\caption{Comparison across different methods on easy document alignment}
\label{tab:alignment_quality_easy}
\footnotesize
\begin{tabular}{llccccc}
\toprule
\multirow{2}{*}{Dataset} & \multirow{2}{*}{Metric} 
& \multicolumn{3}{c}{Baseline approaches} 
& Our approach \\
\cmidrule(lr){3-5} \cmidrule(lr){6-7}
 &  & BERT & BLEU & VEC & GPT  & Gemini  \\
\midrule

\multirow{3}{*}{Easy 1}
 & P     & 0.936 & 0.966 & 0.961 & \textbf{0.987} & \textbf{0.987} \\
 & R     & 0.961 & 0.947 & 0.961 & \textbf{0.993} & \textbf{0.993} \\
 & F$_1$ & 0.948 & 0.957 & 0.961 & \textbf{0.990} & \textbf{0.990} \\

\multirow{3}{*}{Easy 2}
 & P     & 0.985 & 0.919 & 0.985 & \textbf{0.990} & 0.980 \\
 & R     & 0.980 & 0.914 & 0.985 & \textbf{0.995} & 0.990 \\
 & F$_1$ & 0.982 & 0.916 & 0.985 & \textbf{0.993} & 0.985 \\

\multirow{3}{*}{Easy 3}
 & P     & 0.981 & 0.925 & 0.975 & \textbf{1.000} & \textbf{1.000} \\
 & R     & 0.981 & 0.925 & 0.956 & \textbf{1.000} & \textbf{1.000}\\
 & F$_1$ & 0.981 & 0.925 & 0.965 & \textbf{1.000} & \textbf{1.000}\\
\multirow{3}{*}{Easy 4}
 & P     & \textbf{1.000} & 0.944 & 0.970 & 0.980 & 0.990 \\
 & R     & \textbf{1.000} & 0.954 & 0.965 & 0.990 & 0.995 \\
 & F$_1$ & \textbf{1.000} & 0.949 & 0.967 & 0.985 & 0.992 \\
\multirow{3}{*}{Easy 5}
 & P     & \textbf{1.000} & 0.965 & 0.994 & 0.989 & \textbf{1.000} \\
 & R     & \textbf{1.000} & 0.965 & 0.988 & 0.994 & \textbf{1.000}\\
 & F$_1$ & \textbf{1.000} & 0.965 & 0.991 & 0.991 & \textbf{1.000}\\
\midrule
\multirow{3}{*}{Overall}
 & P     & 0.982 & 0.943 & 0.977 & 0.989 & \textbf{0.991}\\
 & R     & 0.985 & 0.943 & 0.972 & 0.994 & \textbf{0.995}\\
 & F$_1$ & 0.984 & 0.942 & 0.974 & 0.992 & \textbf{0.993}\\
\bottomrule
\end{tabular}

\vspace{1mm}
\footnotesize{BERT means BertAlign, BLEU means BlueAlign and VEC means VecAlign. Bold indicates the best-performing method.}
\end{table}

\subsection{Evaluation Metrics}
We use the strict precision, strict recall and strict F1 as our evaluation metrics. Strict true positive, as defined by \cite{sennrich2010mt}, means the alignment hypothesis should be identical to the reference alignment. We adopt the strict metrics because the actual usage of the alignments in translation studies or MT training requires the precise and exact mappings between the source lines and target lines. In this case, the overlapping alignment as lax measures allowed will be considered incorrect. We take the Micro-Averaged Metrics for computing the overall scores for the easy and hard parts.

% In document body:
% \begin{equation}
% \text{Precision} = \frac{\text{TP}}{\text{TP} + \text{FP}}
% \end{equation}

% \begin{equation}
% \text{Recall} = \frac{\text{TP}}{\text{TP} + \text{FN}}
% \end{equation}

% \begin{equation}
% F_1 = 2 \cdot \frac{\text{Precision} \cdot \text{Recall}}{\text{Precision} + \text{Recall}}
% \end{equation}

\section{Experiments \& Results}
\subsection{Experimental settings}
We evaluate the alignment accuracy on the Arabic-English datasets using our proposed LLM-based alignment methods, in comparison with the previous methods, including BertAlign, BleuAlign, and VecAlign. Our experiments are conducted on a computer with Intel Ultra 7 265KF, 64GB RAM, and NVIDIA RTX 5070 Ti. Two LLMs are used in the experiments, including GPT-5.1-mini and Gemini-2.5-flash.
\subsection{Experiments on legal text (easy part)}
An analysis of the experimental results in Table \ref{tab:alignment_quality_easy} suggests that the "Easy" subsets of the dataset may lack the discriminatory power required to fully evaluate the capabilities of aligning methods. As shown by the "Overall" metrics, Gemini achieves a near-perfect $F_1$ score of 0.993; however, the consistently high scores across all methodologies, all exceeding the 0.90 threshold, indicate that these specific benchmarks have reached saturation. Consequently, more challenging data are required to reveal the latent performance gaps.

\begin{table}[htbp]
\setlength{\tabcolsep}{2pt}  % default is 6pt
\centering
\caption{Comparison across different methods on hard document alignment}
\label{tab:alignment_quality_hard}
\footnotesize
\begin{tabular}{llccccc}
\toprule
\multirow{2}{*}{Dataset} & \multirow{2}{*}{Metric} 
& \multicolumn{3}{c}{Baseline approaches} 
& Our approach \\
\cmidrule(lr){3-5} \cmidrule(lr){6-7}
 &  & BERT & BLEU & VEC & GPT  & Gemini  \\
\midrule

\multirow{3}{*}{Hard 1}
 & P     & 0.667 & 0.507 & 0.670 & 0.684 & \textbf{0.811} \\
 & R     & 0.757 & 0.479 & 0.735 & 0.783 & \textbf{0.880} \\
 & F$_1$ & 0.726 & 0.493 & 0.701 & 0.730 & \textbf{0.844} \\

\multirow{3}{*}{Hard 2}
 & P     & 0.724 & 0.563 & 0.791 & 0.718 & \textbf{0.871} \\
 & R     & 0.807 & 0.519 & 0.818 & 0.841 & \textbf{0.920} \\
 & F$_1$ & 0.763 & 0.540 & 0.804 & 0.775 & \textbf{0.895} \\

\multirow{3}{*}{Hard 3}
 & P     & 0.606 & 0.410 & 0.645 & 0.612 & \textbf{0.688} \\
 & R     & 0.648 & 0.410 & 0.682 & 0.682 & \textbf{0.750} \\
 & F$_1$ & 0.626 & 0.410 & 0.663 & 0.645 & \textbf{0.717} \\
 
\multirow{3}{*}{Hard 4}
 & P     & 0.778 & 0.618 & 0.826 & 0.884 & \textbf{0.903} \\
 & R     & 0.865 & 0.547 & 0.854 & \textbf{0.944} & \textbf{0.944} \\
 & F$_1$ & 0.819 & 0.580 & 0.840 & 0.913 & \textbf{0.923} \\
 
\multirow{3}{*}{Hard 5}
 & P     & 0.699 & 0.340 & 0.780 & 0.748 & \textbf{0.880} \\
 & R     & 0.742 & 0.310 & 0.804 & 0.794 & \textbf{0.907} \\
 & F$_1$ & 0.720 & 0.324 & 0.792 & 0.770 & \textbf{0.893} \\
\midrule
\multirow{3}{*}{Overall}
 & P     & 0.696 & 0.494 & 0.743 & 0.729 & \textbf{0.831} \\
 & R     & 0.757 & 0.462 & 0.780 & 0.809 & \textbf{0.831} \\
 & F$_1$ & 0.726 & 0.477 & 0.761 & 0.767 & \textbf{0.855} \\
 
\bottomrule
\end{tabular}

\vspace{1mm}
% \footnotesize{Bold indicates the best-performing method.}
\end{table}

\subsection{Experiments on literary text (hard part)}
% While the preliminary evaluations on simplified datasets suggested a performance parity among various models, 
The results on the challenging document alignment task (Table \ref{tab:alignment_quality_hard}) provide a much clearer differentiation of model capabilities. As the task complexity increases, the "ceiling effect" observed in the easy datasets disappears, revealing significant discrepancies in model robustness. Gemini achieves an overall $F_1$ score of 0.855, which represents a substantial margin over the BertAlign baseline (0.726) and the GPT-based approach (0.767).
The sharp decline in performance across all baseline methods in "Hard" datasets underscores their limitations. For instance, BleuAlign’s overall $F_1$ plummeted from 0.942 to 0.477, and BertAlign’s overall $F_1$ dropped from 0.984 to 0.726. 
More importantly, the divergence between GPT and Gemini on the hard datasets—most notably in Hard 5, where Gemini outperforms GPT by over 12 percentage points in $F_1$—indicates that the selection of LLMs also exerts considerable influence on the alignment quality.

\section{Conclusions}
In this paper, we introduced a generative sentence alignment method and a new Arabic–English parallel dataset focusing on legal and literary texts. Our findings reveal the limitations of previous methods. In contrast, the LLM-based methods  demonstrated better robustness, maintaining high precision and recall where other methods degraded. 

\section*{Limitations}
The proposed LLM-based document alignment methods can effectively and accurately align the low-resource Arabic text to English text. However, the required time for the alignment depends on the processing time of the LLMs. In addition, the format of the response is also essential for the successful parsing and construction of the alignment mappings. Furthermore, the length of the document is also constrained by the context window of the large language models. In our experiments, one simple prompt is used, and the effects of advanced prompt methods, such as few-shot, chain-of-reason (COT), self-reflection, or agentic design, are not explored.

\section*{Acknowledgments}
The authors extend their appreciation to Umm Al-Qura University, Saudi Arabia for funding this research work through grant number: 26UQU4350258GSSR01.

\section*{Funding statement}
This research was funded by Umm Al-Qura University, Saudi Arabia under grant number 26UQU4350258GSSR01.
\bibliography{custom}

\appendix

% \section{Example Appendix}
% \label{sec:appendix}

% This is an appendix.

\end{document}